\documentclass[letterpaper, 10 pt, conference]{ieeeconf}  

\IEEEoverridecommandlockouts                              

\usepackage{tabularx}
\usepackage{multirow}
\usepackage{booktabs}
\usepackage{graphicx}
\usepackage{amsfonts}
\usepackage{subfig}

\usepackage{cite}
\usepackage{amsmath}

\usepackage{varwidth}

\title{\LARGE \bf
Long-term Blood Pressure Prediction with Deep Recurrent Neural Networks
}

\author{Peng~Su, Xiao-Rong~Ding, Yuan-Ting~Zhang,~\IEEEmembership{Fellow,~IEEE,} Jing Liu, Fen~Miao, Ni~Zhao  
    \thanks{Peng Su, Xiao-Rong Ding, Yuan-Ting Zhang, Jing Liu and Ni Zhao are with the Department of Electronic Engineering, The Chinese University of Hong Kong, Hong Kong (psu@ee.cuhk.edu.hk;  nzhao@ee.cuhk.edu.hk). Fen Miao is with the Key Laboratory for Health Informatics of the Chinese Academy of Sciences, Shenzhen Institutes of Advanced Technology, Shenzhen, China.}
    \thanks{Published as a conference paper at IEEE International Conference on Biomedical and Health Informatics (BHI) 2018.}
}

\begin{document}

\maketitle
\thispagestyle{empty}
\pagestyle{empty}


\begin{abstract}
Existing methods for arterial blood pressure (BP) estimation directly map the input physiological signals to output BP values without explicitly modeling the underlying temporal dependencies in BP dynamics. 
As a result, these models suffer from accuracy decay over a long time and thus require frequent calibration. 
In this work, we address this issue by formulating BP estimation as a sequence prediction problem in which both the input and target are temporal sequences. 
We propose a novel deep recurrent neural network (RNN) consisting of multilayered Long Short-Term Memory (LSTM) networks, which are incorporated with (1) a bidirectional structure to access larger-scale context information of input sequence, and (2) residual connections to allow gradients in deep RNN to propagate more effectively.
The proposed deep RNN model was tested on a static BP dataset, and it achieved root mean square error (RMSE) of 3.90 and 2.66 mmHg for systolic BP (SBP) and diastolic BP (DBP) prediction respectively, surpassing the accuracy of traditional BP prediction models.
On a multi-day BP dataset, the deep RNN achieved RMSE of 3.84, 5.25, 5.80 and 5.81 mmHg for the 1st day, 2nd day, 4th day and 6th month after the 1st day SBP prediction, and 1.80, 4.78, 5.0, 5.21 mmHg for corresponding DBP prediction, respectively, which outperforms all previous models with notable improvement.
The experimental results suggest that modeling the temporal dependencies in BP dynamics significantly improves the long-term BP prediction accuracy.
\end{abstract}

\IEEEpeerreviewmaketitle

\section{Introduction}
As the leading risk factor of cardiovascular diseases (CVD) \cite{lim2013comparative}, high blood pressure (BP) has been commonly used as the critical criterion for diagnosing and preventing CVD. 
Therefore, accurate and continuous BP monitoring during people's daily life is  imperative for early detection and intervention of CVD.
Traditional BP measurement devices, e.g., Omron products, are cuff-based and therefore bulky, discomfort to use, and only available for snapshot measurements. These disadvantages restrict the use of the cuff-based devices for 
long-term and continuous BP measurement, which are essential for nighttime monitoring and precise diagnosis of different CVD symptoms.

A key feature of our cardiovascular system is its complex dynamic self-regulation that involves multiple feedback control loops in response to BP variation\cite{guyton1972arterial}. 
This mechanism gives the BP dynamics a temporal dependency nature. Accordingly, such dependency is critical for continuous BP prediction and in particular, for long-term BP prediction.

Existing methods for cuffless and continuous BP estimation can be categorized into two groups, namely physiological model, i.e., pulse transit time model\cite{chen2000continuous} \cite{poon2006cuff}, and regression model, such as decision tree, support vector regression and etc\cite{miao2017novel}\cite{jain2016sparse}. 
These models suffers from accuracy decay over time, especially for multi-day continuous BP prediction. 
Such limitation has become the bottleneck that prevents the use of these models in practical applications. 
It is worth noting that the aforementioned models directly map present input to the target while ignoring the important temporal dependencies in BP dynamics. 
This could be the root of long-term inaccuracy.

Compared with static BP prediction, the multi-day BP prediction is generally much more challenging. 
Due to the complex regulation mechanisms of human body, multi-day BP dynamics have more intricate temporal dependencies and a larger variation range.
In this paper, we formulate the BP prediction as a sequence learning problem, and propose a novel deep RNN model, which is proved to be very effective for modeling long-range dependencies in BP dynamics and
has achieved the state-of-the-art accuracy on multi-day continuous BP prediction.

\section{The Model}
\begin{figure}
\centering
\includegraphics[width=\linewidth]{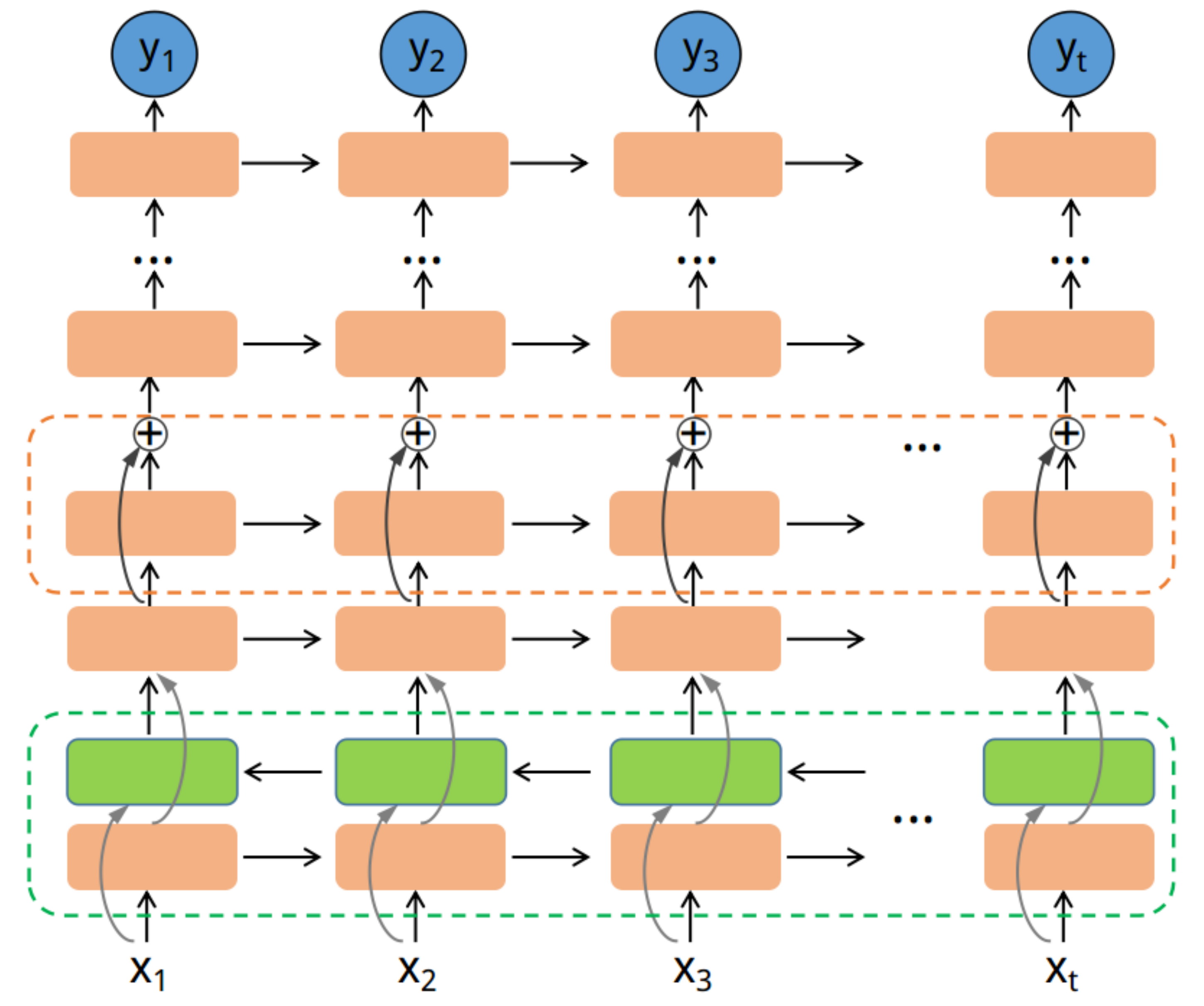}
\caption{DeepRNN architecture. Each rectangular box is an LSTM cell.
The green dashed box at bottom is a bidirectional LSTM layer consisting of forward (orange) and backward (green) LSTM.
The orange dashed box depicts the LSTM layer with residual connections.}
\label{fig:deeprnn}
\end{figure}

The goal of arterial BP prediction is to use multiple temporal physiological signals to predict BP sequence.
Let $ X_T = [ x_1, x_{2} \dots , x_{T}] $ be the input features extracted from electrocardiography (ECG) and photoplethysmogram (PPG) signals, and $ Y_T =  [y_1, y_{2} \dots , y_{T}] $ denote the target BP sequence.
The conditional probability $ p( Y_T \mid  X_T ) $ is factorized as: 
\begin{equation} \label{eq:1}
p( Y_T \mid  X_T ) = \prod_{t =1}^T p(y_{t}\mid h_t),
\end{equation}
where $ h_t $ can be interpreted as hidden state of BP dynamic system and it is generated from previous hidden state $ h_{t-1} $ and current input $ x_t $ as:
\begin{equation} \label{eq:2}
h_t = f(h_{t-1}, x_t).
\end{equation}
Figure \ref{fig:deeprnn} illustrates the overview of our proposed deep RNN model.
The deep RNN consists of a bidirectional LSTM at the bottom layer, and a stack of multilayered Long Short-Term Memory (LSTM) with residual connections.
The full network was trained with backpropagation through time \cite{werbos1990backpropagation}  to miniaturize the difference between BP prediction and the ground truth.

\subsection{Bidirectional LSTM Structure}
First, we introduce the basic block of our deep RNN model, a one-layer bidirectional \textit{Long short-term memory }(LSTM).
LSTM \cite{hochreiter1997long} was designed to address the vanishing gradient problem of conventional RNN by introducing a memory cell state $ c_t $ and multiple gating mechanisms inside a standard RNN hidden state transition process.
The hidden state $h_{t}  $ in LSTM is generated by:
\begin{align}
f_{t} & =  \sigma(W_{xf}x_{t} + W_{hf}h_{t-1} + b_{f}) \label{eq:3} \\
i_{t} & =  \sigma(W_{xi}x_{t} + W_{hi}h_{t-1} + b_{i}) \label{eq:4} \\
o_{t} & =  \sigma(W_{xo}x_{t} + W_{ho}h_{t-1} + b_{o}) \label{eq:5} \\
c_{t} & =  f_{t} \odot c_{t-1}+i_{t}\odot \tanh(W_{xc}x_{t}+W_{hc}h_{t-1}+b_{c}) \label{eq:6} \\
h_{t} & =  o_{t} \odot \tanh(c_{t})  \label{eq:7} 
\end{align}
where $ f$, $i$ and  $o$ are respectively the forget gate, input gate, output gate that control how much information will be forgotten, accumulated, or be outputted.
$ W $ and $ b $ terms denote weight matrices and bias vectors respectively.
$\sigma $ and $ \tanh $ stand for an element-wise application of the logistic sigmoid function and hyperbolic tangent function respectively, and {$\odot $} denote element-wise multiplication.

Conventional LSTMs use $ h_t $ to capture information from the past history $x_{1}, \dots , x_{t-1}  $, and the present input $ x_t $.
To access larger-scale temporal context of input sequence, one can also incorporate nearby future information $ x_{t+1}, \dots , x_{T} $ to inform the downstream modeling process.
Bidirectional RNN (BRNN) \cite{schuster1997bidirectional} can realize this function by processing the data in both forward and backward directions with two separate hidden layers, which then merge to the same output layer. 
As illustrated in the bottom of Figure\ref{fig:deeprnn}, a BRNN computes a forward hidden state $h_t^{f}$, a backward hidden state $h_t^{b} $ and final output $ h_{t} $  by following equations:
\begin{align}
h_{t}^{f} & = \mathcal{H} ( W_{ {h} {h}}^{f}h_{t-1}^{f}+ W_{x h}^{f}x_{t} + b_{f} ) \label{eq:8}\\
h_{t}^{b} & = \mathcal{H} (W_{h h }^{b}h_{t+1}^{b}+ W_{x h}^{b}x_{t} + b_{b})  \label{eq:9} \\
h_{t} & =W^{f}h_{t}^{f} + W^{b}h_{t}^{b} +  b_h \label{eq:10} .
\end{align}
where $ \mathcal{H} $ is implemented by Equations \ref{eq:3}-\ref{eq:7}.

\subsection{Multilayered Architecture with Residual Connections}
A variety of experimental results \cite{graves2013speech}\cite{amodei2016deep} have suggested that RNNs with deep architecture can significantly outperform shallow RNNs.
Simply by stacking multiple layers of RNN could readily gain expressive power. However, a full deep network could become difficult to train as it goes deeper, likely due to exploding and vanishing gradient problems \cite{pascanu2013difficulty}.

\begin{figure}[h]
\centering
\includegraphics[width=0.7\linewidth]{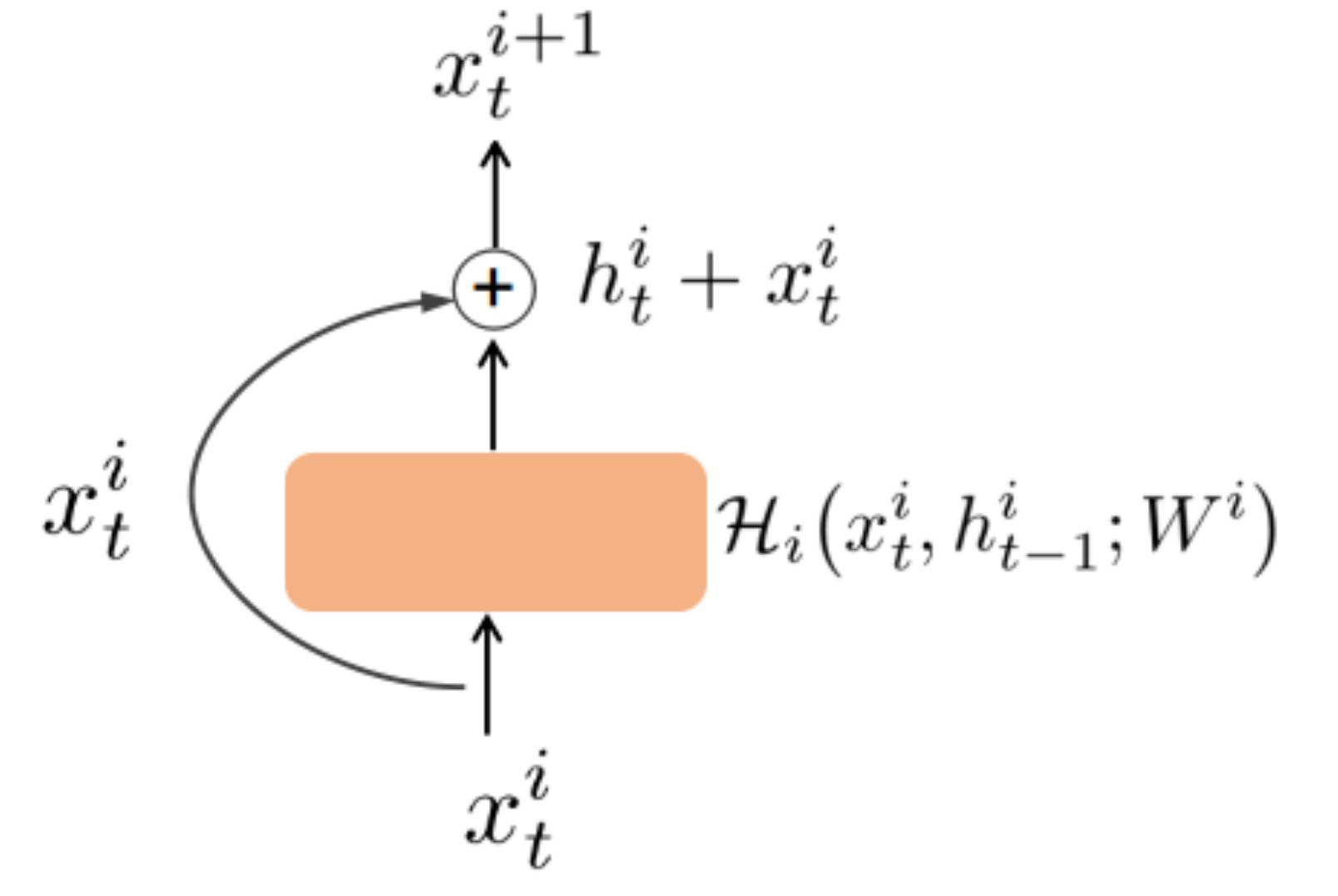}
\caption{LSTM with residual connection.}
\label{fig:residual}
\end{figure}

Inspired by the idea of attaching an identity skip connection between adjacent layers, which
has shown good performance for training deep neural networks\cite{he2016deep}\cite{srivastava2015highway}\cite{wu2016google},
we incorporate a residual connection from one LSTM layer to the next in our model, as shown in Figure \ref{fig:residual}.
Let $x_t^{i} $, $h_t^{i} $, and $\mathcal{H}_i$ be the input, hidden state and LSTM function respectively associated with the $ i$-th LSTM layer ($ i = 1,2,\dots, L $ ), and $W^{i} $ is the corresponding weight of  $\mathcal{H}_i$.
The input to the $ i$-th LSTM layer $ x_t^{i}  $ is element-wise added to this layer's hidden state $ h_{t}^{i} $. This sum $x_t^{i+1}  $ is then fed to the next LSTM layer.
The LSTM block with residual connections can be implemented by:
\begin{align}
h_{t}^{i} & = \mathcal{H}_i \big( x_{t}^{i},  h_{t-1}^{i};  W^{i} \big)  \label{eq:11}\\
x_t^{i+1} & = h_{t}^{i} + x_t^{i} \label{eq:12} \\
h_{t}^{i+1} & = \mathcal{H}_{i+1} \big( x_{t}^{i+1},  h_{t-1}^{i+1};  W^{i+1} \big) \label{eq:13}.
\end{align}
The deep RNN model can be created by stacking multiple such LSTM blocks on top of each other, with the output of previous block  forming the input of the next.
Once the top-layer hidden state is computed, the output $ z_t $ can be obtained by:
\begin{equation}
z_{t}  = \sigma \big( W_{hz}^{L} h_{t}^{L} + W_{xz}^{L} x_{t}^{L} +  b^{L} \big) \label{eq:14} .
\end{equation}

\subsection{Multi-task Training}
Given that we have multiple supervision signals like systolic BP (SBP), diastolic BP (DBP) and mean BP (MBP) which are closely related to each other,
we adopt multi-task training strategy to train one single model to predict SBP, DBP and MBP in parallel. 
Accordingly, the training objective is to minimize the mean squared error (MSE) of total $ N $ training samples as follow:
\begin{equation}
\mathcal{L} ( \{x_{1:T}  ,y_{1:T}\}_N )
	 =  \dfrac{1}{N} \sum_{i=1}^{N} \sum_{t=1}^{T}  {\lVert z_t -y_t \rVert}^2 + \lambda{\lVert  \theta \rVert}^2 \label{eq:15}  ,
\end{equation}
where $y_t=[SBP, DBP, MBP] $ represents ground truth, $ z_t$ is corresponding prediction.
And $ {\lVert  \theta \rVert}^2 $ represents the $ L_2 $ regulation of model parameters and $\lambda  $ is the corresponding penalty coefficient.
One advantage of multi-task training is that learning to predict different BP values simultaneously could implicitly encode the quantitative constrains among SBP, DBP and MBP.

\section{Analysis of Deep RNN Architecture} \label{sec:analysis}
RNNs are inherently deep in time because of their hidden states transition.
Despite the depth in time, the proposed Deep RNN model is also deep along layer structure.
To simplify the analysis, here we mainly focus on the gradient flow along the depth of layers.
Through recursively updating Equation \ref{eq:12}, we will have:
\begin{align}
x_t^{L} & = x_t^{l} + \sum_{i=l}^{L-1} \mathcal{H}_{i} \big( x_{t}^{i},  h_{t-1}^{i};  W^{i} \big)   \label{eq:16} ,
\end{align}
for any deeper layer $ L $ and shallower layer  $ l $.
Equation \ref{eq:16} leads to nice backward propagation properties.
Denoting the loss function as $\mathcal{L}  $, by the chain rule of backpropagation we have:
\begin{align}
\frac{\partial \mathcal{L}}{\partial x_t^{l}} &= \frac{\partial \mathcal{L}}{\partial x_t^{L}} \frac{\partial x_t^{L}}{\partial x_t^{l}}  \nonumber \\
& = \frac{\partial \mathcal{L}}{\partial x_t^{L}} \big( 1 + \frac{\partial}{\partial x_t^{l}} \sum_{i=l}^{L-1} \mathcal{H}_{i} \big( x_{t}^{i},  h_{t-1}^{i};  W^{i} \big)  \big).  \label{eq:17}
\end{align}

Equation \ref{eq:17} shows that the gradient $\frac{\partial \mathcal{L}}{\partial x_t^{l}}  $ can be decomposed into two additive terms: a term of $ \frac{\partial \mathcal{L}}{\partial x_t^{L}} $ that propagates information directly without through any weight layers, and another term $\frac{\partial \mathcal{L}}{\partial x_t^{L}} \big( \frac{\partial}{\partial x_t^{l}} \sum_{i=l}^{L-1} \mathcal{H}_{i} \big) $ that propagates through the weight layers.
The first term of  $ \frac{\partial \mathcal{L}}{\partial x_t^{L}} $  ensures that supervised information could directly backpropagate to any shallower layer $x_t^{l}  $.
In general the term $\frac{\partial}{\partial x_t^{l}} \sum_{i=l}^{L-1} \mathcal{H}_{i}  $ cannot always be $-1$ for all samples in a mini-batch, so the gradient $\frac{\partial \mathcal{L}}{\partial x_t^{l}}  $ is unlikely to be canceled out.
This implies that the gradients of a layer does not vanish even when the intermediate weights are arbitrarily small.
This nice backpropagation property allows us to train deep RNN model that owns more expressive power without worrying about the gradient vanishing problem.

\section{Experiments}
We evaluate the proposed model on both a static and multi-day continuous BP dataset.
Root mean square error (RMSE) is used as the evaluation metric, which is defined as $ RMSE = \sqrt {\dfrac{1}{N} \sum_{i=1}^{N} \sum_{t=1}^{T}\lVert z_t -y_t \rVert ^2} $.
On both datasets we compare our model with the following reference models:
\begin{itemize}
\item
Pulse transit time(PTT) model: we select two most cited PTT-based models - Chen's method \cite{chen2000continuous} \footnote{Chen's PTT model only support SBP prediction, thus only its SBP prediction results were used for comparison with other models.} and Poon's method \cite{poon2006cuff}. 
\item
Typical regression models: support vector regression (SVR), decision tree (DT), and Bayesian linear regression (BLR).
\item
Kalman filter.
\end{itemize}
\subsection{Dataset}
\textbf{Static continuous BP dataset}.
The dataset, including ECG, PPG and BP were obtained from 84 healthy people including
51 males and 33 females.
ECG and PPG signal were acquired with Biopac system and reference continuous BP was measured by Finapres system simultaneously in each experiment.
The BP, ECG and PPG data of each subject were recorded at sampling frequency of 1000 Hz for 10 minutes at the rest status.

\textbf{Multi-day continuous BP dataset}.
Similar dataset was obtained from 12 healthy subjects including 11 males and 1 female.
The BP, ECG and PPG data of each subject were recorded for 8 minutes at the rest status in a multi-day period, namely 1st day, 2nd day, 4th day and 6 moth after the first day.

\subsection{Data Representation}
\begin{figure}
\centering
\includegraphics[width=0.8\linewidth]{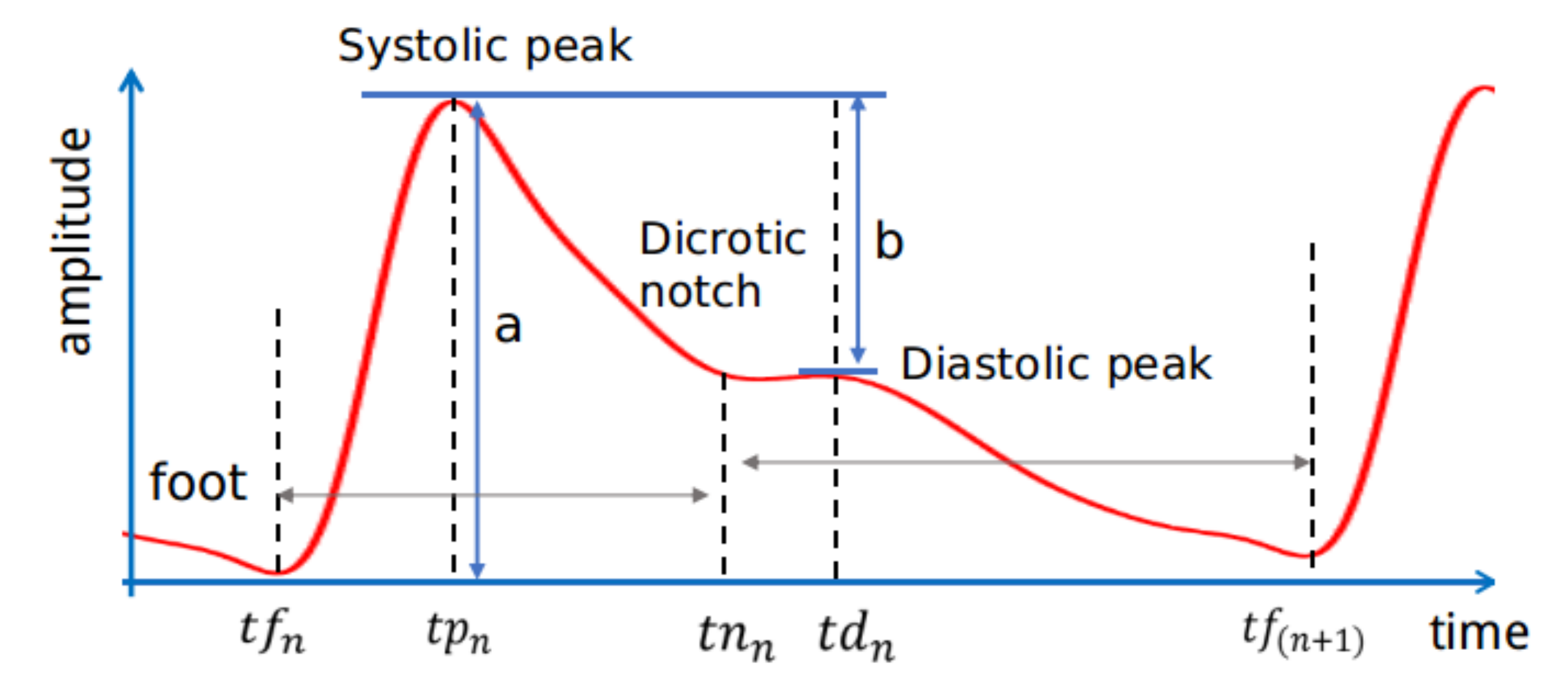}
\caption{Illustration of PPG feature.}
\label{fig:PPG_feature}
\end{figure}

Since the primary goal of this paper is to prove the importance of modeling temporal dependencies in BP dynamics for accurate BP prediction, we simply select 7 representative handcrafted features of ECG and PPG signals (shown in Fig \ref{fig:PPG_feature}) as follows:
\begin{itemize}
\item $ PTT_S $: time interval from ECG R peak to the same heart cycle PPG maximum slope.
\item Heart rate: $ HR $
\item Reflection index: $ RI = b/a $ 
\item Systolic timespan: $ ST = tn_n - tf_n  $ 
\item Up time: $ upTime = tp_n - tf_n   $   
\item Systolic volume: $ SV = \int_{tf_n}^{tn_n} PPG(t) dt  $ 
\item Diastolic volume: $ DV = \int_{tn_n} ^ { tf_{n+1} } PPG(t) dt $ 
\end{itemize}

Now input $X_T$ becomes a $ 7 \times T$ matrix,
and each row of $X_T$ is normalized to have zero-mean and unit-variance. 
Further model performance gain could be expected by adding more informative features as model inputs.

\begin{figure*}
\centering
\includegraphics[width=\linewidth]{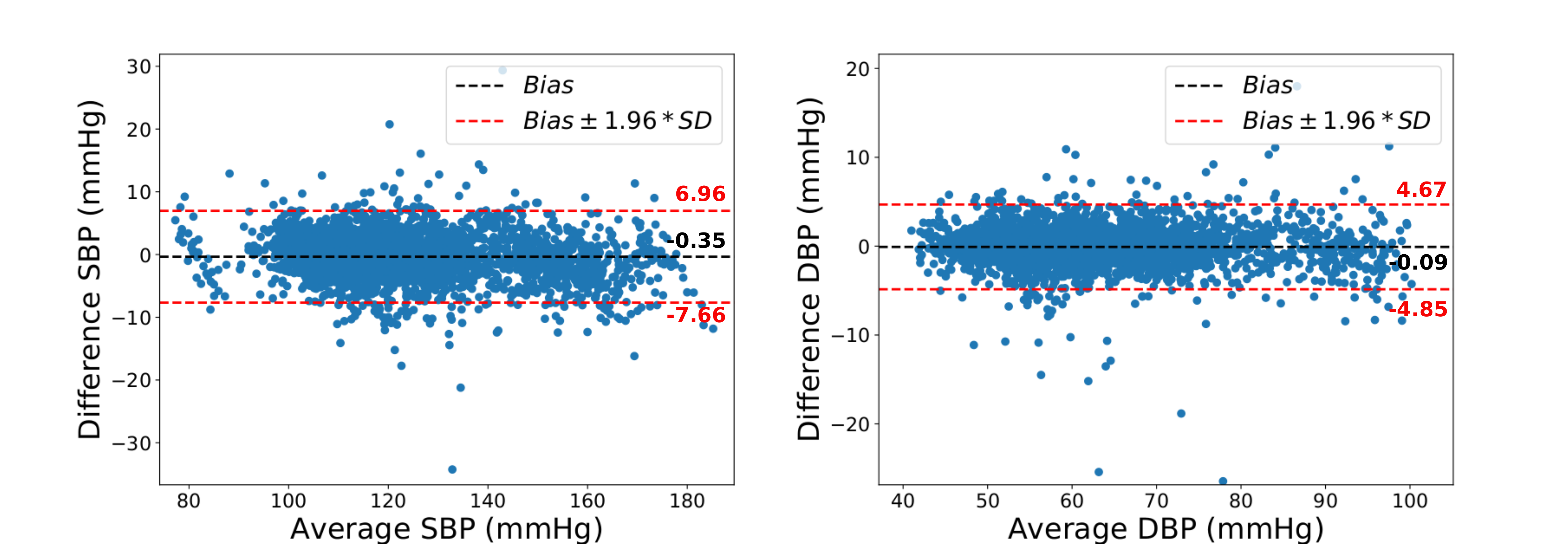}
\caption{Bland-Altman plots of the overall SBP and DBP predictions by a DeepRNN-4L model on the static continuous BP dataset.}
\label{fig:ba_plot}
\end{figure*}

\subsection{Implementation Details}
All the RNN models were trained using mini-batches of size 64 and the Adam optimizer \cite{kingma2014adam}.
For each minibatch, we computed the norm of gradients $ \| g \| $. If  $ \| g \| > v$ , the gradients were scaled by  $ gv / \| g \| $ (v is set as 5 by default.).
We run our model with different number of layers, with hidden state size as 128 at each layer.
The sequence length $ T $ of each training sample is set to 32, and it could be larger if deeper model is adopted.
For saving computational cost, we only adopt bidirectional LSTM at first layer.
Due to limited training samples of our BP prediction problem, 
the maximum depth of deep RNN model was set as 4 to avoid overfitting.
Each training dataset was divided such that 70\% of the data was used for training, 10\% for validation and 20\% for test. 
SBP, DBP and MBP were normalized to $ (0,1] $ by their corresponding maximum, respectively.
For evaluation on the multi-day continuous BP dataset,
all deep RNN models were first pretrained on the static BP dataset then finetuned using part of the first-day data, and finally tested on the rest of the first-day data as well as the following days' data.
\begin{table} 
  \centering  
  \begin{tabularx}{0.7\linewidth}{lcc}
    \toprule
    {\textbf{Model}} & {\textbf{RMSE(SBP)}} & {\textbf{RMSE(DBP)}}  \\ \midrule    \midrule           
    PTT-Chen \cite{chen2000continuous}   	   & 5.31	    & -       \\
    PTT-Poon \cite{poon2006cuff}       &   5.75    &   3.50	   \\   
    \midrule
    BLR      	&  7.45	   	  	& 	6.20  	\\
    SVR  	& 	6.54   	  &  6.28 	 \\
    DT  		&  	4.45	  	&    2.80 	 \\  
    \midrule
    Kalman Filter 	& 5.17  	  	& 3.09   \\
    \midrule
    LSTM 	  &  6.31    	& 4.58 \\
    BiLSTM 	   &  5.25    	  	&  3.04  \\
    DeepRNN-2L   &  5.13	 		& 3.73  \\
    DeepRNN-3L    &  4.92	 	& 3.13 \\
    \textbf{DeepRNN-4L} 
       &  \textbf{3.73}  	& \textbf{2.43}	   \\
  \bottomrule 
  \end{tabularx} 
   \caption{Detailed analysis of our Deep RNN models with comparison with different reference models. DeepRNN-xL represents a x layer RNN model. All the models are validated on the static continuous BP dataset. (unit: mmHg)}
   \label{tab:1dayNEW}
\end{table}

\begin{figure*}
    \centering
    \subfloat{{\includegraphics[width=0.48\linewidth]{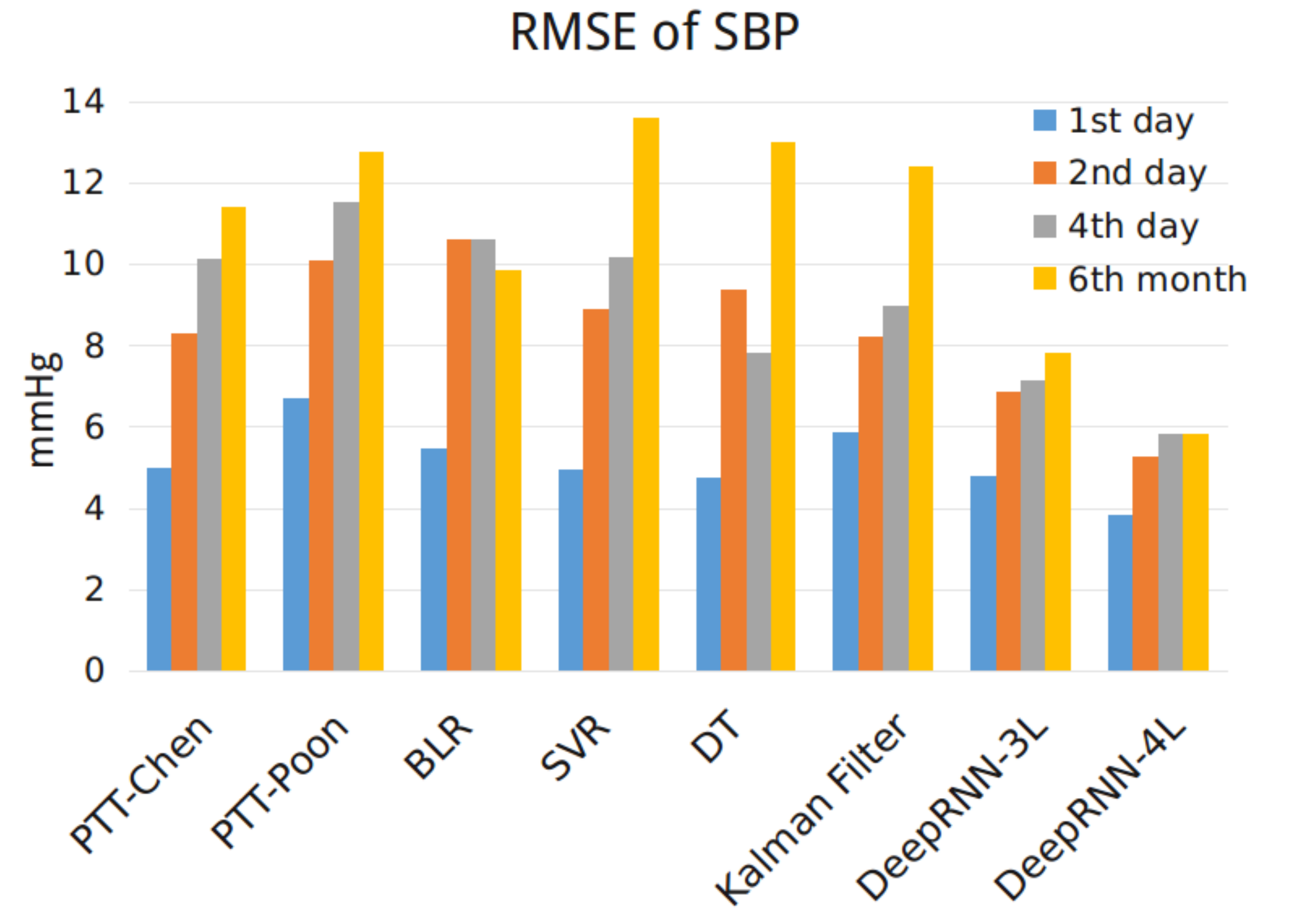} }}%
    \quad
    \subfloat{{\includegraphics[width=0.48\linewidth]{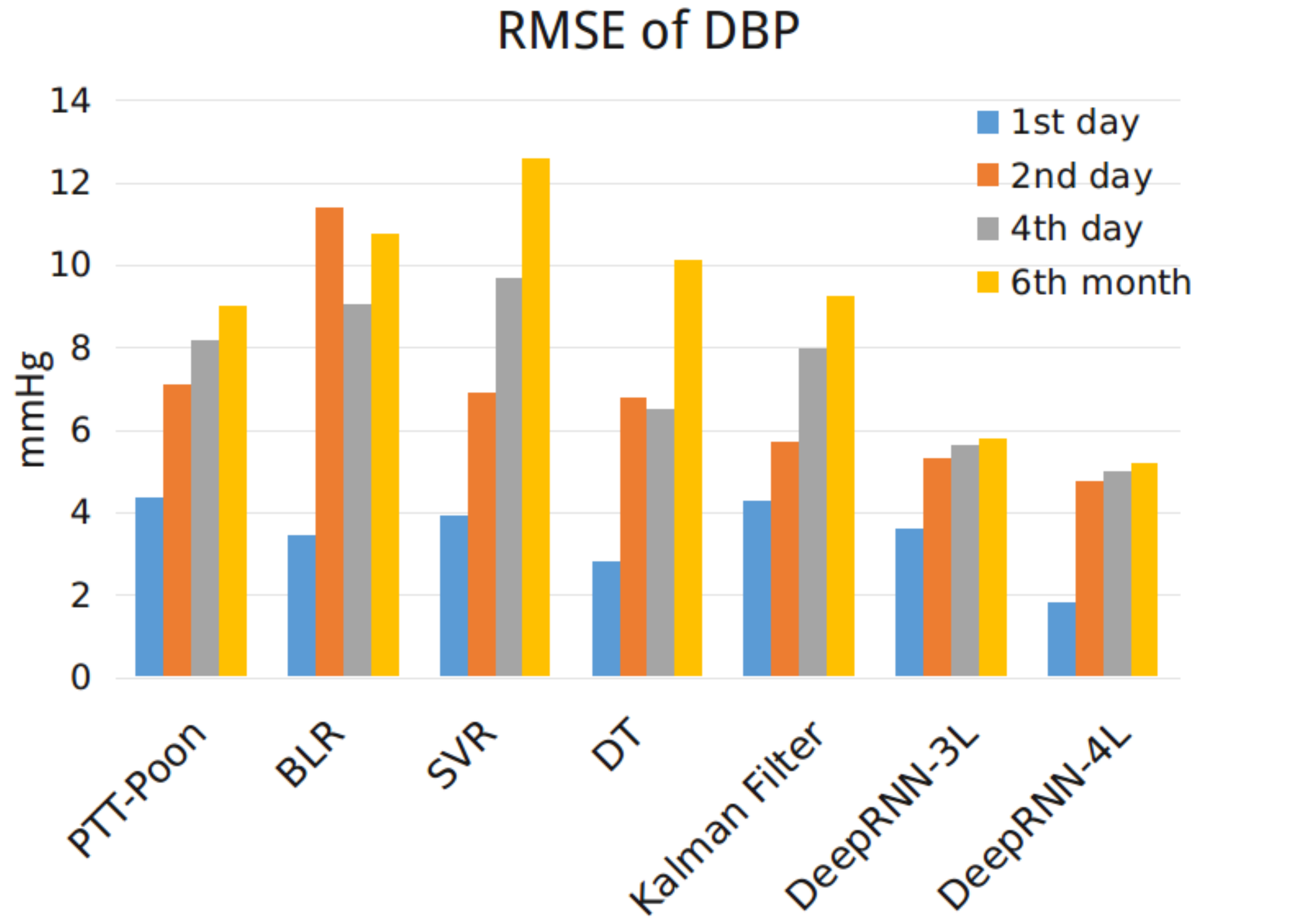} }}%
    \caption{Overall RMSE comparison of different models on the multi-day continuous BP dataset.}
    \label{fig:4day_rmse}
\end{figure*}

\section{Experimental Results}
\begin{figure*}
\centering
\includegraphics[width=0.96\linewidth]{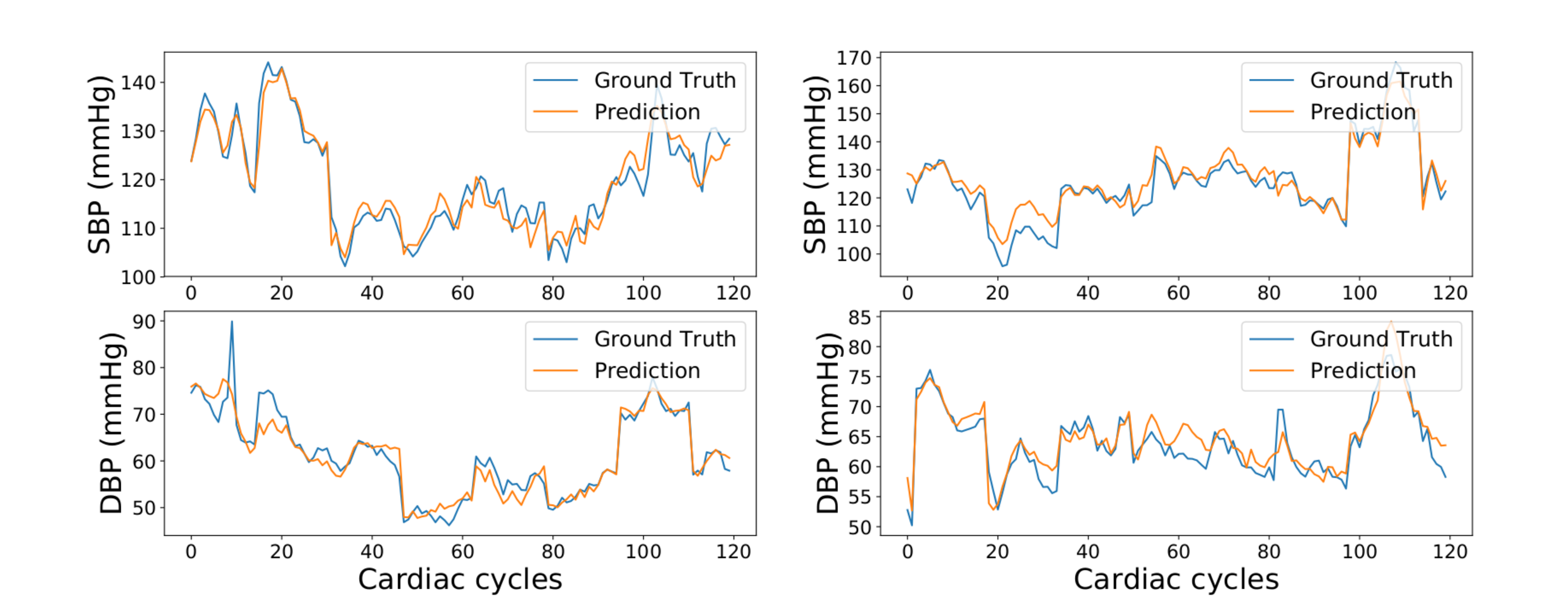}
\caption{Comparison of the ground truth and DeepRNN prediction of two  representative subjects from static continuous BP dataset. Each column represents the SBP and DBP predictions of one subject.}
\label{fig:1day}
\end{figure*}

\textbf{Validation on static continuous BP dataset.}
As shown in Table \ref{tab:1dayNEW}, the PTT models yield slightly better results than BLR and SVR models, but show poorer performance than DT, kalman filter, bidirectional LSTM and deep RNN (DeepRNN) models.
The best accuracy was obtained by our 4-layer deep RNN (DeepRNN-4L) model which achieves a RMSE of 3.73 and 2.43 for SBP and DBP prediction respectively.
The Bland-Altman plots (Figure \ref{fig:ba_plot}) indicate that the DeepRNN-4L predictions agreed well with the ground truth, with 95\% of the differences lie within the agreement area.
Figure \ref{fig:1day} qualitatively shows the DeepRNN-4L prediction result on a representative subject from the static continuous BP dataset.

By incorporating a bidirectional structure in the model,
i.e. by using the bidirectional LSTM (BiLSTM), the
prediction accuracy is improved significantly as compared to the vanilla LSTM, with 17 \% decrease in the SBP RMSE and 34 \% decrease in DBP RMSE.  
Furthermore, it was observed that the improvement of prediction accuracy is enhanced with increasing depth of the DeepRNN network.
For instance, replacing DeepRNN-2L with DeepRNN-4L results in 27\% and 35\% improvement on SBP and DBP prediction respectively.
When we stack up to a 5-layer DeepRNN, the model tend to overfit and no clear benefits of depth can be observed any more.

\begin{figure*}
\centering
\includegraphics[width=\linewidth]{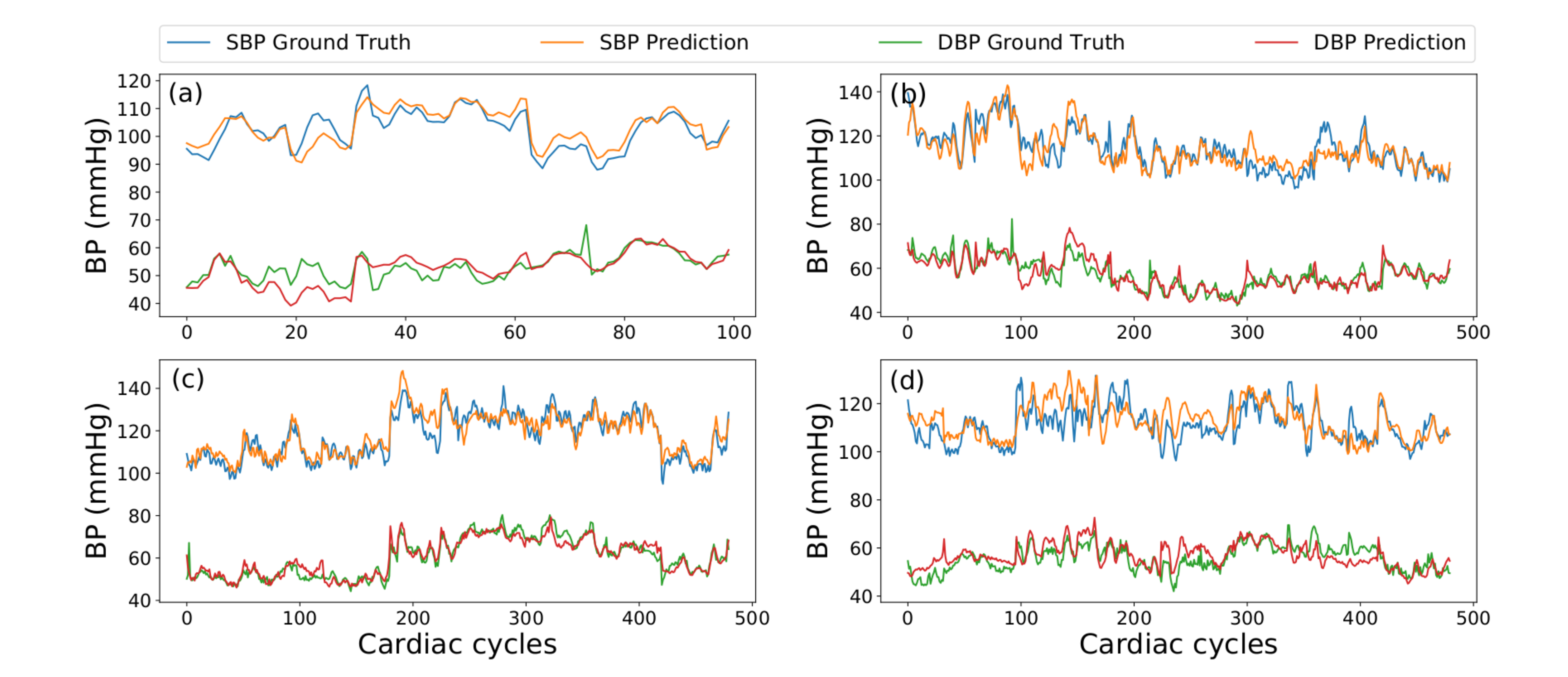}
\caption{Comparison of the ground truth and DeepRNN prediction of one representative subject from multi-day continuous BP dataset. Figure (a), (b), (c) and (d) represent the results of 1st day, 2nd day, 4th day and 6th month after the 1st day, respectively.}
\label{fig:2}
\end{figure*}

\textbf{Validation on multi-day continuous BP dataset.}
Figure \ref{fig:4day_rmse} compares the prediction performance of deep RNN against the reference models.
It can be clearly seen that the DeepRNN models yield much better performance as compared to the PTT and regression models, likely due to the temporal dependencies modeling in the DeepRNN models.
Kalman filter could model the time dependencies in sequence but dose not perform as well as DeepRNN models.
It is likely because of the linearity assumption of kalman filter that both state transition and measurement functions are linear.
This assumption may limit its capability to model the complex temporal dependencies in BP dynamics.
The best accuracy was obtained by our DeepRNN-4L model which achieves a RMSE of 3.84, 5.25, 5.80 and 5.81 mmHg for the 1st day, 2nd day, 4th day and 6th month after the 1st day SBP prediction, and 1.80, 4.78, 5.0, 5.21 mmHg for corresponding DBP prediction, respectively.
As shown in Figure \ref{fig:4day_rmse}, all the PTT models, regression models and kalman filter exhibit pronounced accuracy decay from the second day. 
Although the prediction accuracy of the DeepRNN model also drops after the first day, it consistently provides the lowest RMSE values among all models.
Figure \ref{fig:2} qualitatively shows the capability of DeepRNN to track long-term BP variation.

\textbf{Importance of residual connections.}
To investigate the importance of residual connections, we conduct ablation study on the static continuous BP dataset.
As shown in Table \ref{tab:residual}, DeepRNN model incorporated with residual connections works considerably better than the counterpart.
During training, we found residual connections significantly improve the gradient flow in the backward pass which make deep neural network easier to optimize.
Accordingly, better performance could be obtained due to more expressive deep structure.
The detailed reason for such computational benefit has been explained in section \ref{sec:analysis}.
 
 \begin{table}[]
 \centering
 \begin{tabular}{r|c|c}
 \hline
     		 & RMSE (SBP) & RMSE (DBP) \\ \hline \hline
 DeepRNN-4L w/o residual & 5.31          & 3.13       \\ \hline
 DeepRNN-4L w residual  & 3.73          & 2.43        \\ \hline
 \end{tabular}
 \caption{ Performance comparison of DeepRNN-4L model with residual connections \textit{vs.} without residual connections.
 The result is obtained on the static continuous BP dataset. (unit: mmHg)}
 \label{tab:residual}
 \end{table}
 
 \begin{table}[]
 \centering
 \begin{tabular}{r|c|c}
 \hline
     		 & RMSE (SBP) & RMSE (DBP) \\ \hline \hline
 DeepRNN-2L  & 6.24          & 4.55       \\ \hline
 DeepRNN-3L  & 5.05         & 3.30         \\ \hline
 DeepRNN-4L  & 4.27         & 3.02          \\ \hline 
 \hline
 DeepRNN-2L \text{\dag}  & 5.13          & 3.73      \\ \hline
 DeepRNN-3L \text{\dag}  &  4.92	 	& 3.13         \\ \hline
 DeepRNN-4L \text{\dag} & 3.73          & 2.43          \\ \hline 
 \end{tabular}
 \caption{Investigation of DeepRNN with different settings.
 Models trained using multi-task objective are marked with '$\dag$' .
  The result is obtained on the static continuous BP dataset. (unit: mmHg)}
 \label{tab:multitask}
 \end{table}
 
\textbf{Importance of multi-task training.}
Table \ref{tab:multitask} shows that multi-task training strategy can boost the prediction performance as compared with separate training of individual models.
It can be explained by that the different training objectives involved in each task are strongly correlated and thus share a lot of data representations that capture 
the underlying factors, which can be learned by the same model structure.
Hence, by learning the shared representations, it can crucially improve the model generalization ability.

\section{Conclusions}
In this work, we demonstrated that modeling the temporal dependency in BP dynamics can significantly improve long-term BP prediction accuracy, which is one of the most challenging problems in cuffless BP estimation.
We proposed a novel deep RNN that incorporated with bidirectional LSTM and residual connections to tackle this challenge.
The experimental results show that the deep RNN model
achieves the state-of-the-art accuracy on both static and multi-day continuous BP datasets.

\bibliographystyle{IEEEtran}
\bibliography{jay}

\end{document}